\documentclass[10pt,twocolumn,letterpaper]{article}

\usepackage[pagenumbers]{cvpr} 

\usepackage{graphicx}
\usepackage{amsmath}
\usepackage{amssymb}
\usepackage{booktabs}
\usepackage{pifont}
\usepackage{appendix}
 \usepackage{array,multirow}
 \usepackage{float}

\usepackage{microtype}
\usepackage{booktabs} 
\usepackage{multirow}
\usepackage{caption}
\usepackage{bbm}
\usepackage{setspace}
\usepackage[ruled,vlined]{algorithm2e}
\usepackage{xcolor}

\def\eg{\emph{e.g}.}

\newcommand{\xmark}{\ding{55}}%

\DeclareMathOperator*{\argmin}{arg\,min}

\usepackage[pagebackref,breaklinks,colorlinks]{hyperref}

\newcommand{\smallsec}[1]{\vspace{0.2em}\noindent\textbf{#1}}

\usepackage[capitalize]{cleveref}
\crefname{section}{Sec.}{Secs.}
\Crefname{section}{Section}{Sections}
\Crefname{table}{Table}{Tables}
\crefname{table}{Tab.}{Tabs.}

\def\cvprPaperID{3978} 
\def\confName{CVPR}
\def\confYear{2022}

\begin{document}

\title{Discovering Objects that Can Move}
\author{
{Zhipeng Bao\thanks{Equal contribution} \textsuperscript{,}\thanks{Work done during an internship at TRI} $^{, 1}$ \qquad Pavel Tokmakov\footnotemark[1] $^{, 2}$ \qquad Allan Jabri$^{3}$} \\ {Yu-Xiong Wang$^{4}$ \qquad Adrien Gaidon$^{2}$ \qquad Martial Hebert$^1$} \\
{ $^1$CMU \qquad $^2$Toyota Research Institute \qquad $^3$ UC Berkeley \qquad $^4$ UIUC}\\
}
\maketitle

\begin{abstract}
 This paper studies the problem of object discovery -- separating objects from the background without manual labels. Existing approaches utilize appearance cues, such as color, texture, and location, to group pixels into object-like regions.
 However, by relying on appearance alone, these methods fail to separate objects from the background in cluttered scenes. 
 This is a fundamental limitation since the definition of an object is inherently ambiguous and context-dependent. To resolve this ambiguity, we choose to focus on dynamic objects -- entities that can move independently in the world.
 We then scale the recent auto-encoder based frameworks for unsupervised object discovery from toy synthetic images to complex real-world scenes.  To this end, we simplify their architecture, and
 augment the resulting model with a weak learning signal from general motion segmentation algorithms. Our experiments demonstrate that, despite only capturing a small subset of the objects that move, this signal is enough to generalize to segment both moving and static instances of dynamic objects. 
 We show that our model scales to a newly collected, photo-realistic synthetic dataset with street driving scenarios. Additionally, we leverage ground truth segmentation and flow annotations in this dataset for thorough ablation and evaluation.
 Finally, our experiments on the real-world KITTI benchmark demonstrate that the proposed approach outperforms both heuristic- and learning-based methods by capitalizing on motion cues. 
\end{abstract}

 
\section{Introduction}
Objects are the key building blocks of perception~\cite{kahneman1992reviewing,spelke2007core}. 
We understand the world not in terms of pixels, surfaces, or entire scenes, but rather in terms of individual objects and their combinations. Object-centric representation makes tractable higher-level cognitive abilities such as casual reasoning, planning, etc., and is crucial for generalization and adaptation~\cite{berner2019dota,vinyals2019grandmaster}. In computer vision, progress has been achieved in object recognition recently~\cite{ren2015faster,he2017mask,carion2020end}, but these approaches rely on large amounts of expensive manual labels, and only cover a fixed vocabulary of object categories. Discovering objects and their extent in data -- in a manner that generalizes across domains -- remains an open problem.

\begin{figure}[t]
    \centering
    \includegraphics[width = 1 \linewidth]{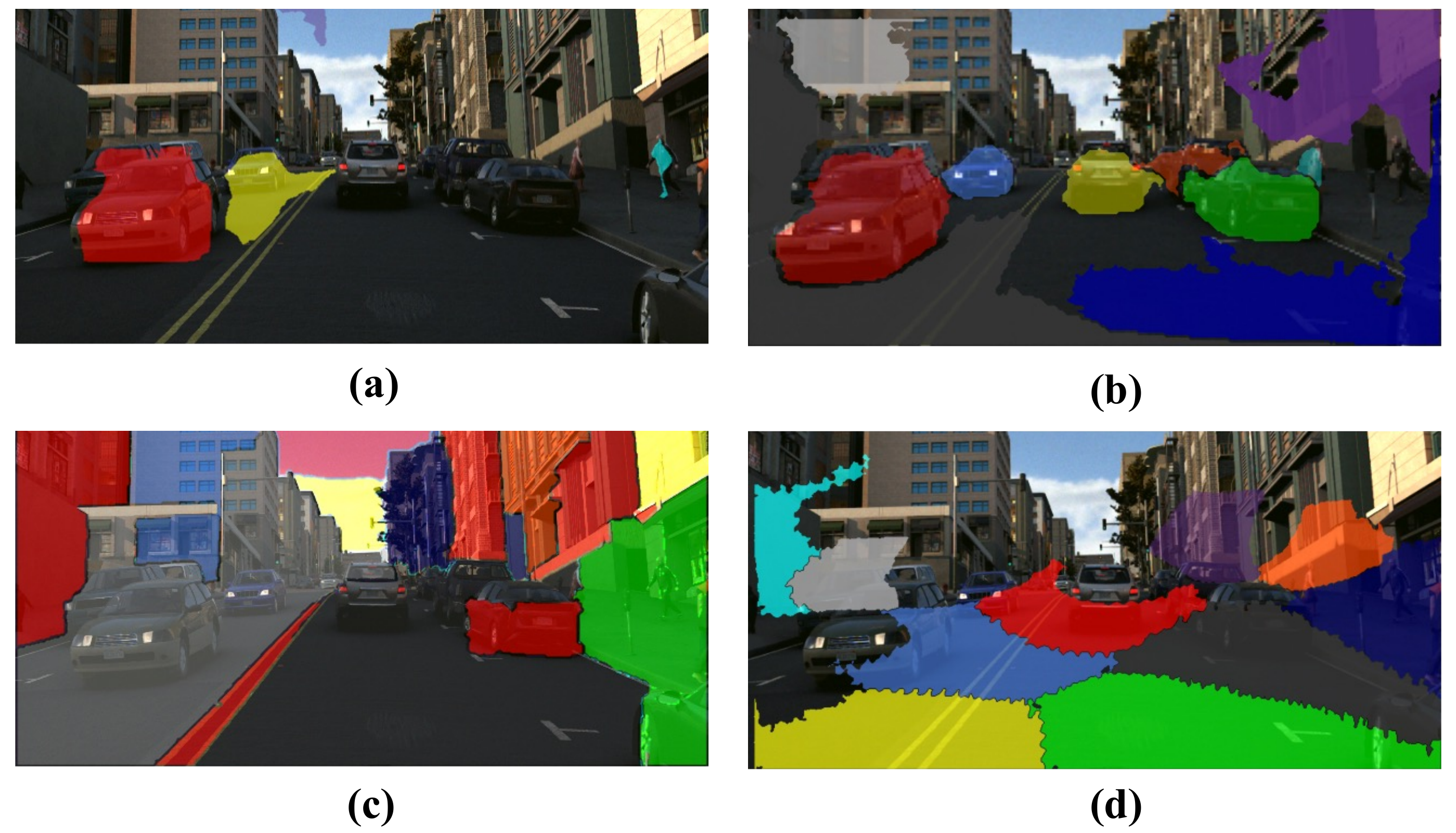}
    \vspace{-5mm}
    \caption{
    A sample from the TRI-PD dataset with: (a) motion segmentation from~\cite{dave2019towards}, top-10 segments produced by (b) our approach, (c) heuristic-based MCG~\cite{arbelaez2014multiscale}, and (d) learning-based SlotAttention~\cite{locatello2020object}. Our method uses noisy, sparse motion segmentation to learn to separate both moving and static instances of dynamic objects from the background, whereas others cannot resolve the object definition ambiguity based on appearance alone.
    }
    \vspace*{-4mm}
    \label{fig:teaser}
\end{figure}

What makes this task especially challenging is that the notion of an object is inherently ambiguous and context-dependent. Consider a car in Figure~\ref{fig:teaser}: its left door and the handle on that door can be treated as individual objects, or parts of the whole. It is thus not surprising that existing approaches that attempt to automatically separate objects from the background based on appearance struggle beyond controlled scenarios. In particular, classical methods that use graph-based inference tend to over- or under-segment the objects~\cite{felzenszwalb2004efficient,arbelaez2014multiscale} (Figure~\ref{fig:teaser}, bottom left). More recent learning-based methods model object discovery with structured generative networks, often leveraging iterative inference in the bottleneck of an auto-encoder~\cite{burgess2019monet,greff2019multi,engelcke2019genesis,lin2020space,locatello2020object}. 
While promising results have been demonstrated, these approaches are typically restricted to toy images with colored geometric shapes on a plain background, and completely fail on realistic scenes (Figure~\ref{fig:teaser}, bottom right).

We posit that while the ambiguity of the object definition is not resolvable in the static image world without direct supervision, it has a natural resolution in the dynamic world of videos. Concretely, we choose to focus on \textit{dynamic} objects, which we define as entities that are capable of moving independently in the world.
Independent object motion is a strong grouping cue, which has been shown to drive object learning in animal perception~\cite{Cynader1353, spelke1990principles}. In computer vision, there exists a long line of works on motion segmentation that automatically separate moving objects from the background based on optical flow~\cite{brox2010object,ochs2012higher,ochs2012higher,keuper2015motion,dave2019towards,xie2019object}. These methods have found numerous applications in unsupervised~\cite{agrawal2015learning,pathak2017learning} and weakly-supervised machine learning algorithms~\cite{prest2012learning,tokmakov2016weakly,hong2017weakly}.

In this work, we show how motion segmentation can be bootstrapped to group instances even when they are static.
We build our approach on top of the framework for unsupervised object discovery proposed by Locatello et al.~\cite{locatello2020object}, and show how to scale it from toy images to realistic videos. We extend the architecture to videos of arbitrary length by introducing a spatio-temporal memory module~\cite{ballas2015delving}, and simplify the grouping mechanism to scale the model to realistic scenes with large resolution and dozens of objects. We then demonstrate the importance of inductive biases based on independent object motion on the emergent representation and the extent to which it captures objects.
In particular, we show how motion segments (Figure~\ref{fig:teaser}, top left) can guide the attention operation to discover static objects. Crucially, we show that motion segmentation of varying quality -- even when sparse and noisy -- can be sufficient to bias the model towards segmenting \textit{both moving and static instances} (Figure~\ref{fig:teaser}, top right). Our approach only requires videos for training, and can segment objects in static images at inference time. 

To go beyond the toy data used in~\cite{locatello2020object}, while still being able to thoroughly analyze the various aspects of the method, we utilize a new, photo-realistic, synthetic dataset collected using the ParallelDomain platform~\cite{parallel_domain}  (TRI-PD). It consists of hundreds of videos, with crowded, street driving scenes, and comes with a full set of ground truth annotations, including object segmentations, 3D coordinates and optical flow, allowing us to ablate the importance of the quality of the motion segmentation to the method's performance. Finally, we demonstrate that the resulting method generalizes to real videos on the challenging KITTI dataset~\cite{geiger2012we}, and compare it to existing heuristic- and learning-based approaches. Our code, models, and synthetic data are made available at \url{https://github.com/zpbao/Discovery_Obj_Move/}.

\section{Related work}
In this work we study the problem of \textit{object discovery} in realistic videos capitalizing on \textit{motion segmentation} as a \textit{learning signal for bottom-up grouping}. Below, we review the most relevant works in each of these areas.

\smallsec{Object discovery} is the problem of separating objects from the background without manual labels. Traditional computer vision approaches treated it as perceptual grouping~\cite{koffka2013principles} -- the idea that low and mid-level regularities in the data such as color, orientation, and texture allow for approximately parsing a scene into object-like regions. Notable approaches include~\cite{felzenszwalb2004efficient}, which uses graph-based inference to identify region boundaries, and~\cite{arbelaez2014multiscale} which first extracts regions on multiple scales with a normalized cut algorithm, and then groups them into object candidates. However, being purely appearance-based, these methods are not well equipped to resolve the inherent ambiguity of the object definition.

This problem has received renewed attention recently with the introduction of learning-based methods for object discovery~\cite{eslami2016attend,greff2016tagger,burgess2019monet,greff2019multi,engelcke2019genesis,lin2020space,locatello2020object,veerapaneni2020entity,jiang2019scalor,yu2021unsupervised}. A common approach is to use iterative inference to bind a set of variables to objects in an image~\cite{greff2019multi,engelcke2019genesis,locatello2020object}, usually with a variational auto-encoder~\cite{kingma2013auto,rezende2014stochastic}. A more efficient variant is proposed by Locatello et al.~\cite{locatello2020object} in their SlotAttention framework. Concretely, they perform a single step of image encoding with a CNN (convolutional neural network) followed by an iterative attention operation, which is used to bind a set of variables, called slots, to image locations. The slots are then decoded individually and combined to reconstruct the image. 

Many of the approaches above are capable of discovering objects in toy, synthetic scenes, but as we demonstrate in Section~\ref{sec:sota}, they fail in more realistic environments, where appearance alone is not sufficient to separate the objects from the background. In this work, we extend SlotAttention to realistic videos by modifying the architecture of the model to allow it to scale to large scenes with dozens of objects, and incorporating inductive biases in the form of motion segmentation. Crucially, our method only uses motion as a sparse learning signal and the trained model is able to segment both moving and static instances.

Finally, several works have recently explored integrating inductive biases in the form of 3D geometry constraints~\cite{stelzner2021decomposing,chen2020object,du2020unsupervised,henderson2020unsupervised}. However, these methods remain limited to toy, synthetic environments. In contrast, our method uses independent object motion as a learning signal, allowing it to generalize to real-world scenes. Geometric priors are orthogonal to our approach and combining different forms of inductive biases is a promising direction for future work.

\smallsec{Motion segmentation} is concerned with separating objects from the background using optical flow~\cite{ilg2017flownet,teed2020raft,stone2021smurf}. Early approaches~\cite{brox2010object,ochs2012higher,ochs2012higher,keuper2015motion} tracked individual pixels with the flow and then clustered the resulting trajectories inspired by the common fate principle~\cite{koffka2013principles}. While these methods have shown promising results on motion segmentation benchmarks, they do not generalize well in the wild due to their heuristic-based nature. More recently, several learning-based methods have been proposed~\cite{dave2019towards,xie2019object}. In particular, Dave et al. re-purpose a state-of-the-art object detection architecture~\cite{he2017mask} to detect and segment moving objects in an optical flow field. The model is trained on a toy, synthetic FlyingThings3D dataset~\cite{mayer2016large}, but can generalize to real videos due to appearance abstraction provided by the flow. We use this method in our work due to its high performance and simplicity combined with minimal supervision requirements. Note that since our method requires instance-level moving object masks, binary motion segmentation techniques~\cite{yang2021self,tokmakov2019learning,papazoglou2013fast} are not applicable in our scenario.

\smallsec{Learning from motion} is a paradigm inspired by evidence from cognitive science research, that independent object motion is a crucial cue for the development of the human visual system~\cite{spelke1990principles}. In computer vision, it has been adopted for weakly-supervised object detection~\cite{prest2012learning} and semantic segmentation~\cite{tokmakov2016weakly,hong2017weakly}, as well as for unsupervised representation learning~\cite{pathak2017learning,agrawal2015learning}. However, none of these works address the problem of object discovery from unlabeled videos. Yang et al.~\cite{yang2021dystab} use binary motion segmentation to train saliency models, but do not segment individual objects in complex scenes.  Very recently, Tangemann et al.~\cite{tangemann2021unsupervised} have proposed to use motion segmentation to build compositional, generative scene models. However, their approach employs motion segmentation as a pre-processing step during training and is not capable of object discovery at inference time.

\section{Method}
\begin{figure*}[t]
    \centering
    \includegraphics[width = 0.95 \linewidth]{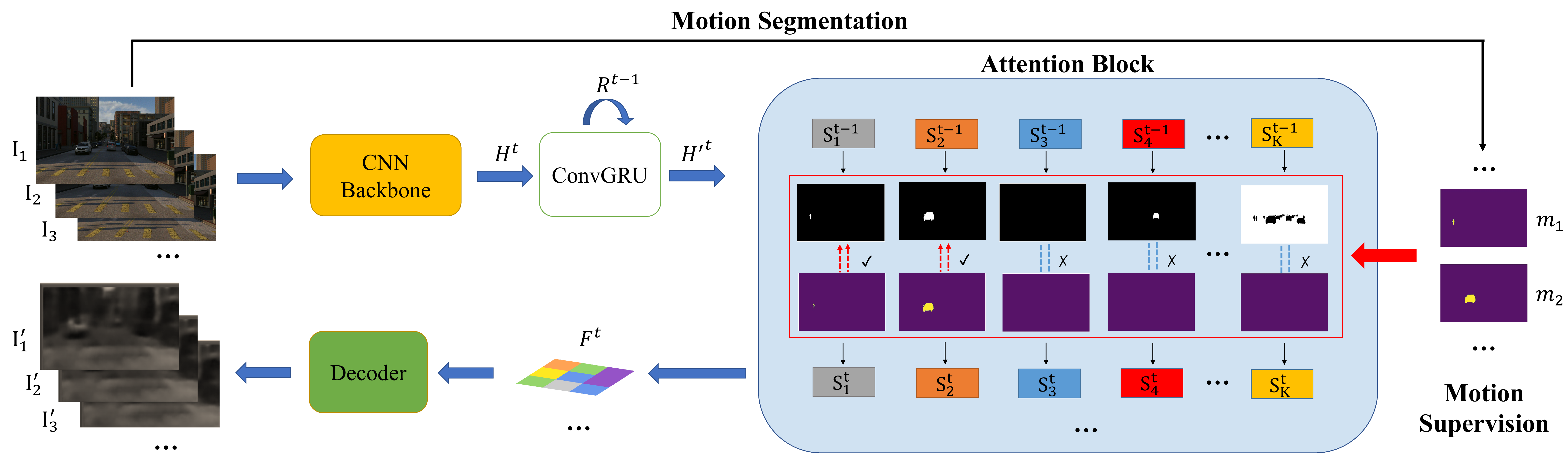}
    \vspace{-2 mm}
    \caption{Our method takes a sequence of frames as input and processes them individually with a backbone network (shown in yellow), and a ConvGRU recurrent memory module. The resulting feature maps $H^{'t}$ are passed to the attention module (shown in blue) which binds them to a fixed set of slot variables via an attention operation. We additionally use automatically estimated motion segmentation to guide the attention operation for a subset of the slots. Finally, the slot states are combined in a single feature map $F^t$ and decoded to reconstruct the frame. The reconstruction objective enforces generalization from moving to static instances.}
     \vspace*{-4mm}
    \label{fig:model}
\end{figure*}

In this section, we first introduce the SlotAttention framework for unsupervised object discovery, which serves as a basis for our approach, in Section~\ref{sec:bckg}. We then describe how we scale this architecture to real-world videos with dozens of objects in Section~\ref{sec:scale}, and present our approach to incorporating independent motion priors in Section~\ref{sec:prior}.

\subsection{Background}
\label{sec:bckg}
Following prior work~\cite{greff2019multi, burgess2019monet}, SlotAttention~\cite{locatello2020object} models object discovery as inference in an auto-encoder framework. Concretely, given an image $I \in \mathbb{R}^{H \times W \times 3}$, it is first passed through an encoder CNN to obtain a hidden representation $H = f_{enc}(I) \in \mathbb{R}^{H' \times W' \times D_{inp}}$. It is then processed by the attention module, which we describe below, to map $H$ to a set of feature vectors of a fixed length $K$ called slots $S \in \mathbb{R}^{K \times D_{slot}}$. Each slot $S_i\in S$ is broadcasted onto a 2D grid, and decoded individually with a decoder CNN $O_i = f_{dec}(S_i) \in \mathbb{R}^{H \times W \times 4}$, where the 4th dimension of the output represents the alpha mask $A_i$. Denoting the first 3 channels of $O_i$ with $I^{'}_i$, the complete image reconstruction is obtained via $I' = \sum_i A_i * I^{'}_i$ and is used to supervise the model with an MSE (mean squared error) loss.

The attention module is the key component of the approach. It uses an iterative attention mechanism, similar to the one used in Transformer~\cite{vaswani2017attention}, to map from the input $H$ to the slots $S$. In particular, the attention weights are computed with a dot product between the input features and slot states $W = \frac{1}{\sqrt{D}} k(H) \cdot q(S) \in \mathbb{R}^{N \times K}$, where $k$ and $q$ are learnable linear transformations and $N = H' \times W'$.  These attention weights are then used to compute the update values via $U = W^T v(H) \in \mathbb{R}^{K \times D}$, where $W$ are the normalized attention weights, and $v$ is another linear transformation. A key difference to the classical Transformer architecture is that the slots are initialized at random, and the inference is iterative. In particular, at every step $l$ the slots are updated via $S_l = {\tt update}(S_{l - 1}, U_l)$, where the update function is implemented as a GRU~\cite{cho2014learning} (gated recurrent unit).
 
The intuition behind this approach is that the slots serve as a representational bottleneck and individual decoding of the slots results in them binding to spatially coherent regions, such as objects. Next, we describe how we modify the SlotAttention framework to scale it to real-world videos.
 
\subsection{A framework for object discovery in videos}
\label{sec:scale}
Our model, shown in Figure~\ref{fig:model}, takes a sequence of video frames $\{I^{1},I^2, ..., I^T\}$ as input. Following~\cite{locatello2020object}, each frame is then processed by an encoder CNN, shown in yellow, to obtain an individual frame representation $H^t = f_{enc}(I^t)$. These individual representations are aggregated by a ConvGRU spatio-temporal memory module~\cite{ballas2015delving} to obtain video encoding via $H^{'t}={\tt ConvGRU}(R^{t-1}, H^t)$, where $R^{t-1} \in \mathbb{R}^{H' \times W' \times D_{inp}}$ is the recurrent memory state. 

Next, we proceed to map the video representation $H^{'t}$ to the set of slots $S^t$. It is easy to see, however, that the recurrent slot assignment strategy proposed in~\cite{locatello2020object} does not scale well to sequential inputs. Indeed, given a sequence of length $T$ and $L$ inference steps for each frame, the overall number of attention operations required to process the sequence is $T \times L$. Such a nested recurrence is both computationally inefficient, and can exacerbate the vanishing gradient problem. To address this issue, as shown in the blue block in Figure~\ref{fig:model}, we only perform a single attention operation to directly compute the slot state $S^t = W^{t^T} v(H^{'t})$, where the attention matrix $W^t$ is computed using the slot state in the previous frame $S^{t-1}$. For the first frame we use a learnable initial state $S^0$.

It is worth noting that the authors of~\cite{locatello2020object} suggest that iterative inference on randomly initialized slots is crucial for the model to be able to generalize to a different number of objects at test time. However, we have found that simply increasing the number of slots to the maximum expected number of objects is sufficient to  generalize to scenes of varying complexity. In that regard, our approach is similar to DETR~\cite{carion2020end}, which also uses transformer query vectors as learnable object proposals that are capable of parsing both densely and sparsely populated scenes, but is trained in a fully supervised way.

Finally, the resulting slot states $S^t$ need to be processed with the decoder CNN, shown in green in Figure~\ref{fig:model}, to obtain the frame reconstruction. However, the individual slot decoding approach from~\cite{locatello2020object} does not scale well with the number of slots. Indeed, a full image reconstruction needs to be computed for each slot which quickly becomes prohibitively expensive in terms of memory, especially for large resolution frames. Instead, we propose to invert the order of slot decoding and slot recombination steps. In particular, we first broadcast each individual slot feature $S_i^t \in  \mathbb{R}^D_{slot}$ to a feature map $F_i^t \in \mathbb{R}^{H' \times W' \times D_{slot}}$ and use the attention mask $W_{:,i}^t$ of the slot as an alpha mask $A_i^t$. We then construct a single output feature map $F^t=\sum_i A_i^t * F_i^t$, shown with a checkerboard pattern in the figure, and decode it via $I^{'t} = f_{dec}(F_t) \in \mathbb{R}^{H \times W \times 3}$.

As we demonstrate in Section~\ref{sec:arch}, the proposed single shot decoding strategy reduces the strength of the spatial cohesion prior to the original SlotAttention architecture, decreasing its object discovery capabilities. However, we also demonstrate that this prior does not generalize beyond toy, synthetic scenes. Instead, in the next section we describe our approach of incorporating an independent motion prior which provides a stronger learning signal and works well with a single shot decoding strategy.

\subsection{Incorporating independent motion priors}
\label{sec:prior}
Our method assumes that a set of sparse, instance-level motion segmentation masks $\mathcal{M} = \{M^1, M^2, ..., M^T\}$ is provided with every video, with $M^t = \{m_1, m_2, ..., m_{C^t}\}$, where $C^t$ is the number of moving objects that were successfully segmented in frame $t$, and $m_j \in \{0,1\}^{H' \times W'}$ is a binary mask (downsampled to match the spatial dimension of the feature maps). Note that for every frame it is possible that $M_i = \emptyset$. This reflects the realistic assumption that a variable number of objects can be moving in any given frame and that in some frames all the objects can be static.

We propose to use these motion segmentation masks to directly supervise the slot attention maps $W^t \in \mathbb{R}^{N \times K}$. We thus need to map a variable number of motion segmentations $C^t$ to a fixed number of slots $K$ in every frame. Following prior work on set-based supervision~\cite{carion2020end,stewart2016end}, we first find an optimal bipartite matching between predicted and motion masks, and then optimize an object-specific segmentation loss. Specifically, we consider $M^t$ also as a set of length $K$ padded with $\emptyset$ (no object). To find a bipartite matching between these two sets we search for a permutation of $K$ elements with the lowest cost: 
\begin{equation}
    \hat{\sigma} = \argmin_{\sigma} \sum_{i=1}^K \mathcal{L}_{seg}(m^t_i, W^t_{:,\sigma(i)}),
\end{equation}
where $\mathcal{L}_{seg}(m^t_i, W^t_{:,\sigma(i)})$ is the segmentation loss between the motion mask $m^t_i$ and the attention map of the slot with index $\sigma(i)$. In practice, we efficiently approximate the optimal assignment with a greedy matching algorithm.

Once the assignment $\hat{\sigma}$ has been computed, the final motion supervision objective is defined as follows:
\begin{equation}
    \mathcal{L}_{motion} = \sum_{i=1}^K  \mathbbm{1}_{\{m^t_i \neq \emptyset\}} \mathcal{L}_{seg}(m^t_i, W^t_{:,\hat{\sigma}(i)}).
\label{eq:motion}
\end{equation}
That is, the loss is only computed for the slots for which motion masks have been assigned, and the remaining slots are not constrained and can bind to any regions in the image. This is illustrated in the right part of Figure~\ref{fig:model}, where motion segmentation masks are available for only two objects in a crowded outdoor scene, and they get matched to the slots whose attention maps are most similar to the masks. The remaining slots are unconstrained, but still manage to capture both moving and static objects, as well as the background, driven by the image reconstruction objective. The actual segmentation loss $\mathcal{L}_{seg}$ in Eq.~\ref{eq:motion} is the binary cross entropy:
\begin{equation} 
\small 
\mathcal{L}_{seg}(m, W) 
=  \sum_{j=1}^N - m_j \log (W_j) 
- (1 - m_j) \log (1 - W_j).
\end{equation}

\subsection{Loss function and optimization}
\label{sec:loss}
Our final objective is defined as follows:
\begin{equation}
    \mathcal{L} = \mathcal{L}_{recon} + \lambda_{M} \mathcal{L}_{motion} + \lambda_{T} \mathcal{L}_{temp},
\end{equation}
where $\mathcal{L}_{recon}$ is the MSE loss for the image reconstruction, $\mathcal{L}_{temp}$ is a temporal consistency regularization term, and $\lambda_M$ and $\lambda_T$ are the weights for the motion supervision and temporal consistency terms. The latter is defined as 
\begin{equation}
    \mathcal{L}_{temp} (S) = \sum_{t=1}^{T-1} \| \mathbb{I} - softmax (S^t \cdot (S^{t+1})^{\bf{T}}) \| ,
\end{equation}
where  $\mathbb{I} \in \mathbb{R}^{K \times K}$ is the identity matrix. It is easy to see that this term is encouraging similarity between feature representations of the slots in consecutive frames and thus improving temporal consistency of the slot bindings. The model is trained on video clips of length $T$ and we ensure that at least half of the clips in a batch have a non-empty set of motion segmentations $\mathcal{M}$.
\section{Experimental evaluation}

\subsection{Datasets and evaluation}

We use two synthetic datasets for the analysis of the proposed approach: CATER~\cite{girdhar2019cater} for ablating the architecture of the model and a realistic ParallelDomain (TRI-PD) dataset for analyzing the impact of the motion segmentation quality on the model's performance. In addition, we use a real-world KITTI benchmark~\cite{geiger2012we} for comparison to the state of the art.

\begin{figure}[t]
    \centering
    \includegraphics[width = 0.95 \linewidth]{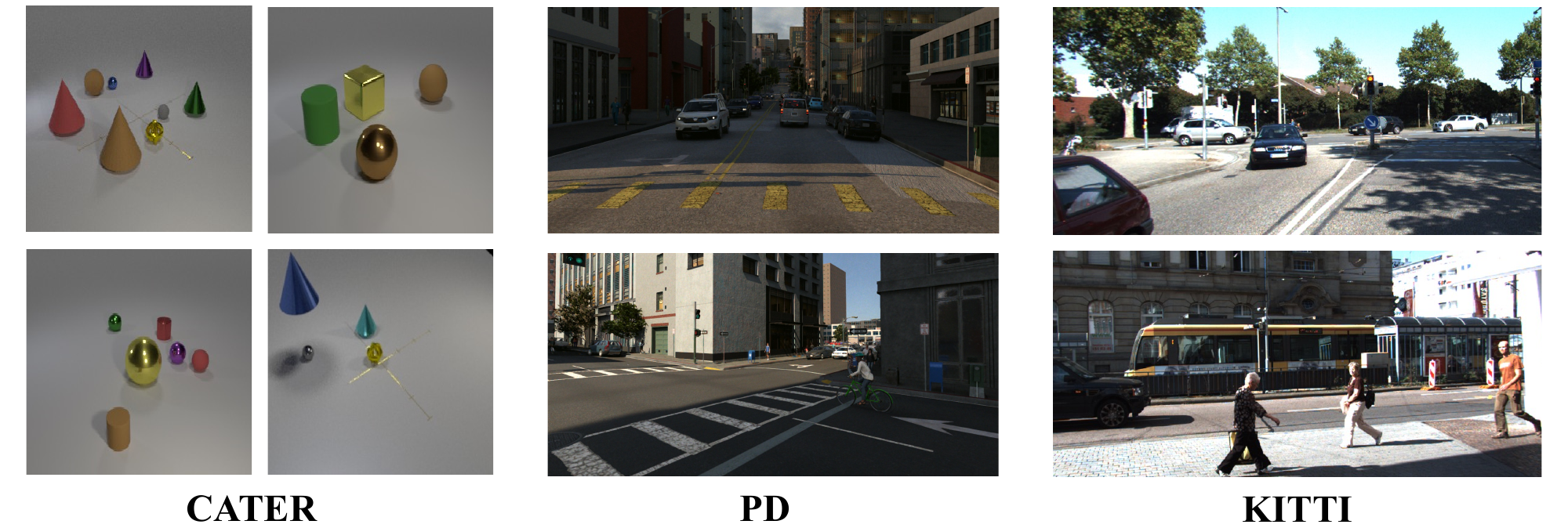}
    \vspace{-2 mm}
    \caption{Frame samples from the video datasets used in our experiments. CATER~\cite{girdhar2019cater} (left) is a toy, synthetic dataset similar to the ones used in prior works. TRI-PD (center) is a collection of photo-realistic, synthetic videos, which is a major step forward in visual complexity. KITTI~\cite{geiger2012we} (right) is a real world benchmark with outdoor scenes.}
    \vspace*{-4mm}
    \label{fig:data}
\end{figure}

\smallsec{CATER} is a video version of the CLEVR~\cite{johnson2017clevr} dataset which was used in many recent works on unsupervised object discovery~\cite{locatello2020object,burgess2019monet,jiang2019scalor}.  We utilize the provided engine to generate 2,000 videos by placing between 4 and 8 geometric shapes, such as cubes or cones, on a plain background at random, and assigning a random color to each instance. Each object can then move on a random trajectory or remain static, and the camera motion is also randomized. We use 1600 videos for training and 400 for evaluation, with each video being 40-frames long with a resolution $128 \times 128$ (see Figure~\ref{fig:data}, left). 
For ablation analysis, we randomly assign one object as moving in each video and use the ground truth mask of that object as a motion mask. Notice that we do experiment with automatically estimated motion segmentation on more challenging TRI-PD and KITTI.

\smallsec{ParallelDomain (TRI-PD)} is a synthetic dataset with street driving scenarios (see Figure~\ref{fig:data}, center). It was collected using a state-of-the-art synthetic data generation service~\cite{parallel_domain}. The training set contains 924 photo-realistic, 10 seconds long videos with driving scenarios in city environments captured at 20 FPS. We use 51 videos from a disjoint set of scenes for evaluation. Each video comes with a full set of ground truth annotations, including optical flow, allowing us to conduct a detailed analysis of the impact of the motion segmentation quality on our method's performance. More statistics and qualitative examples are provided in the appendix. 

\smallsec{KITTI} is a real-world benchmark with city driving scenarios which comes with a variety of annotations (Figure~\ref{fig:data}, right). In this work, we use the instance segmentation subset of the dataset for evaluation. It contains 200 frames, which we resize to $368 \times 1248$. Notice that instance segmentation annotations are provided on individual images in this dataset, without the temporal context, allowing us to demonstrate that our approach does not require videos at inference time. Since our model is unsupervised, we use all the 147 videos in the training set of KITTI to discover the objects that can move in the real world.

\smallsec{Evaluation metrics.} We use Adjusted Rand Index (ARI) as the main metric for comparing object discovery capabilities of the models, but also report more traditional segmentation metrics like F-measure and mIoU in the appendix. ARI is a clustering similarity metric which captures how well predicted segmentation masks match ground-truth masks in a permutation-invariant fashion. This is more suitable for the evaluation of unsupervised approaches than, say, mIoU, because it does not require for the methods to make the decision which segments represent the objects and which correspond to the background. Following prior work~\cite{locatello2020object,greff2019multi}, we only measure ARI based on foreground objects, which we refer to as Fg. ARI.

\subsection{Implementation details}
For the components of our model shared with SlotAttention~\cite{locatello2020object} we follow their architecture and training protocol exactly, and describe the remaining details below. 

We replace the shallow encoder used in~\cite{locatello2020object} with a ResNet18~\cite{he2016identity} to scale the representational power to realistic scenes. We also experiment with deeper backbones in the appendix. All the models are trained from scratch unless stated otherwise. We additionally report results with contrastive-learning pre-training in the appendix. To be able to capture small objects, we remove the last 2 max pooling layers from the ResNet, and add a corresponding dilation ratio to preserve the field of view. We use 10 slots for the experiments on CATER and 45 slots on TRI-PD and KITTI to account for the larger number of objects.

All the models are trained for 500 epochs using Adam~\cite{kingma2014adam} with a batch size 20 and learning rate 0.001. Following~\cite{locatello2020object}, we use learning rate warm-up~\cite{goyal2017accurate} and an exponential decay schedule to prevent early saturation and reduce variance. We set $\lambda_M$ to 0.5 and $\lambda_T$ to 0.01 on the validation set of CATER, and use these value in all the experiments. Video-based variants are trained using clips of length 5. At inference time, the model is evaluated in a sliding window fashion with a stride 5.

We experiment with two motion segmentation algorithms -- a heuristic-based~\cite{keuper2015motion}, and a learning-based one~\cite{dave2019towards}, for which we only use the motion stream trained on the toy FliyingThings3D dataset~\cite{mayer2016large}. Both methods take optical flow as input, so we evaluate them with both ground truth flow, and flow estimated with the state-of-the-art supervised~\cite{teed2020raft} and unsupervised~\cite{stone2021smurf} approaches. Since the outputs of both methods contain many noisy segments, we apply a few generic post-processing steps to clean up the results. They remove very large and very small segments, as well as segments at the image boundary. The details of the post-processing are provided in the appendix. 

We compare our approach to several recent learning-based object discovery algorithms as well as to a classical, heuristic-based method. In particular, we choose SlotAttention~\cite{locatello2020object}, MONet~\cite{burgess2019monet}, SCALOR~\cite{jiang2019scalor}, and S-IODINE~\cite{greff2019multi} as a representative sample of learning-based methods, with S-IODINE also being a video-based approach. For MONet and S-IODINE, we replace the original backbone with ResNet18 and match the input resolution to the one used by our method for a fair comparison, but keep all the other details intact. All the models are trained until convergence. We use MCG~\cite{arbelaez2014multiscale} as an heuristic-based baseline. It is a proposal generation method, so to obtain a single interpretation of an image, we sample the top scoring proposals until all the pixels are covered. For overlapping segments, we assign the corresponding pixels to the smaller segment.

\subsection{Architectural analysis}
\label{sec:arch}
In this section, we begin the analysis of our method by studying the variants of the auto-encoder framework for object discovery on the validation set of CATER in Table~\ref{tab:arch_anal}. Firstly, we evaluate the original SlotAttention model (row 1 in the table), which serves as a basis for our approach, and find that it performs reasonably well on this toy dataset, though the Fg. ARI scores are noticeable lower than those reported in the original paper~\cite{locatello2020object} on CLEVR. This is explained by the fact that the scenes in CATER are more challenging, with a larger variance in the number of objects and more occlusions. 
\begin{table}[bt]
 \centering
  {
\resizebox{\linewidth}{!}{
\begin{tabular}{c|c@{\hspace{1em}}|c@{\hspace{1em}}|c@{\hspace{1em}}|c@{\hspace{1em}}|c@{\hspace{1em}}|c@{\hspace{1em}}}
    ConvGRU & Slot inf. & Temp. & Decode. & Motion & Recon. & Fg. ARI        \\\hline
    -- & Iter. & \xmark & Per slot & \xmark & \checkmark  & 64.4   \\
    \hline
     frame & Iter. & \xmark & Per slot & \xmark & \checkmark  & 66.3  \\
     clip & Iter. & \xmark & Per slot & \xmark & \checkmark  & 71.5   \\
     clip & 1-shot & \xmark & Per slot & \xmark & \checkmark  &   83.2 \\
     clip & 1-shot &\checkmark  & Per slot & \xmark  & \checkmark  &  86.7 \\
    clip & 1-shot & \checkmark  & 1-shot & \xmark  & \checkmark  & 34.5  \\
    clip & 1-shot & \checkmark  & 1-shot &\checkmark & \checkmark   &  {\bf 92.7}  \\
   clip & 1-shot & \checkmark  & 1-shot &\checkmark & \xmark   &   77.9  \\
\end{tabular}
}
}
\caption{Analysis of the model architecture using Fg.~ARI on the validation set of CATER. We ablate the ConvGRU module, slot inference strategy, temporal consistency constraint, decoding strategy, independent motion prior, and the reconstruction objective. Combining motion priors with reconstruction leads to best results.}
\vspace{-4mm}
\label{tab:arch_anal}
\end{table}

Next, we convert the frame-level architecture of SlotAttention to a video-level model by adding a ConvGRU after the encoder. This has only a minor effect on the performance when trained on 1-frame sequences (row 2 in the table), but training on video clips (row 3) results in a 5.2 points increase in Fg.~ARI score. This demonstrates that the feature space of the recurrent model can capture video dynamics and thus simplify separating objects from the background.

However, going from single frame inputs to clips increases the memory requirements of the model. To mitigate this issue, we now study the architectural modifications proposed in Section~\ref{sec:scale}. Firstly, replacing iterative inference on randomly initialized slots with a single attention operation with a learnable initialization not only results in an improved computational efficiency, but also significantly improves the performance. Incorporating the temporal consistency term in the loss further boosts the Fg.~ARI score due to more robust slot binding. Next, switching to 1-shot decoding significantly reduces the memory consumption of the model, but also results in it largely loosing its object-discovery capabilities. This demonstrates that individual slot decoding was crucial for enforcing the spatial cohesion prior to the SlotAttention model. 

Despite this disadvantage, incorporating a weak learning signal in the form of a motion segmentation not only recovers, but significantly improves the model's performance. This demonstrates that independent motion is a much stronger and more generic prior than appearance and location similarity used in the SlotAttention, even in a toy dataset like CATER. Finally, the last row of Table~\ref{tab:arch_anal} shows that the reconstruction objective is still important for achieving top performance by enforcing generalization from moving to static instances. 

\subsection{Object discovery in realistic videos}
\label{sec:disc}
\begin{figure*}[t]
    \centering
    \includegraphics[width = 0.99 \linewidth]{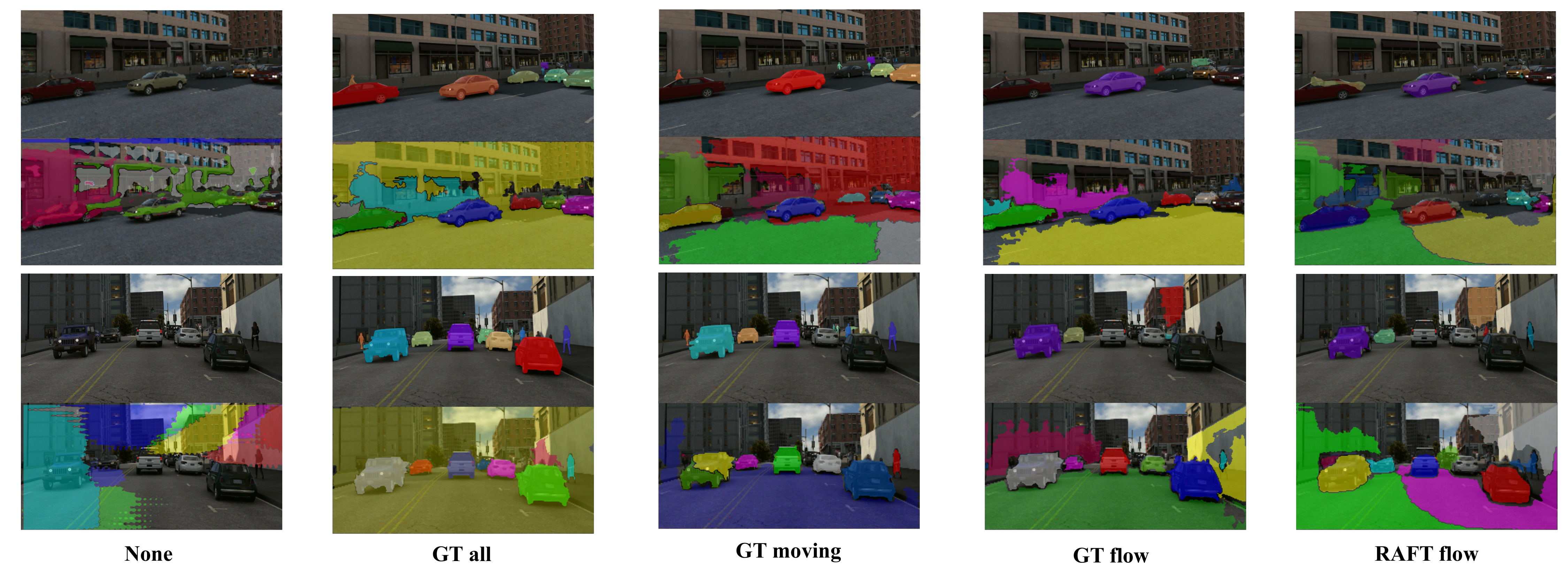}
    \caption{Top-10 masks produced by our model with varying quality of motion priors on the validation set of TRI-PD. We show the motion masks used for supervision on top of the corresponding model's outputs. In the last two columns the approach of Dave et al.~\cite{dave2019towards} is used for motion segmentation. Our method learns to discover the objects even with sparse and noisy motion segmentation based on estimated flow.}
    \vspace{-3mm} 
    \label{fig:pd}
\end{figure*}

We now explore how well the model introduced above scales to realistic outdoor scenes in the TRI-PD dataset in Table~\ref{tab:motion_anal} and Figure~\ref{fig:pd}. We separately report the Fg.~ARI score for moving and static objects to assess the network's generalization abilities.  We begin with evaluating the baseline variant of our model without independent motion priors, and observe that appearance similarity is indeed not sufficient for object discovery in realistic scenes, as reflected by the low Fg.~ARI score. Qualitatively, the first column of Figure~\ref{fig:pd} illustrates that this variant completely fails to discover any objects, and instead segments the scene into random patches based on color and location similarity.

\begin{table}[bt]
 \centering
  {
\resizebox{\linewidth}{!}{
    \begin{tabular}{l|l|c@{\hspace{1em}}|c@{\hspace{1em}}|c@{\hspace{1em}}}
    Model & Motion seg. & Fg. ARI Stat. & Fg. ARI Mov. & Fg. ARI All            \\\hline
    Ours & None & 10.5 & 18.4 & 13.1  \\\hline
    Ours & GT all & 69.0 & 72.2 &  {\bf 71.7} \\
    Ours & GT moving & 53.3 & 62.7 & 59.6 \\
    Ours & GT flow + \cite{keuper2015motion}  & 39.9 &  47.5 &  42.8      \\
    Ours  & GT flow + \cite{dave2019towards} & 48.3  & 54.9 &  51.7    \\
    Ours & RAFT flow + \cite{dave2019towards}  & 46.8 & 55.6 & 50.9  \\
    Ours & SMURF flow + \cite{dave2019towards}  & 47.3 & 54.8 & 50.5  \\\hline
    - & RAFT flow + \cite{dave2019towards}  & 2.7 & 5.3 & 3.4 
\end{tabular}
}
}
\caption{Analysis of the effect of the quality of motion segmentation on the model's performance on validation set of TRI-PD. We gradually reduce the quality of the motion segments starting from ground truth to fully estimated. Our method learns to discover both moving and static instances guided by a very sparse motion signal.}

\vspace{-3mm}
\label{tab:motion_anal}
\end{table}

Next, we establish the upper bound for our model's performance by using all the ground truth object segments (corresponding to moving and static objects) for training. This fully-supervised approach reaches a Fg.~ARI score of 71.7, which is significantly below the 92.7 obtained by the best version of our model on CATER, further emphasizing the complexity of TRI-PD. Qualitatively, as can be seen in the second column of Figure~\ref{fig:pd}, this variant successfully captures all the clearly visible objects in a scene, and also groups the background pixels together. 

Only using the ground truth segments corresponding to the moving objects, which simulates the theoretical scenario in which we have a perfect motion segmentation algorithm, does result in a performance drop of 11.3 Fg.~ARI points, which is especially noticeable for static objects, but the overall score remains 46.5 points higher than the baselines trained without the motion prior. Qualitatively, the model is able to accurately segment most of the moving and static instances, as shown in the third column in Figure~\ref{fig:pd}. However, this variant oversegments the background, demonstrating that explaining as many objects in the scene as possible is crucial for learning a strong background model.

Switching to actual motion segmentation algorithms, we first compare the state-of-the-art heuristic-based and learning-based methods using the ground truth optical flow as input in rows 5 and 6 of the Table~\ref{tab:motion_anal}. As expected, we observe that the more recent learning-based method produces more accurate motion segmentations, which in turn results in a higher performance of our approach. Qualitatively, this model, shown in the 4th column in Figure~\ref{fig:pd}, has a slightly lower recall than the variant trained with ground truth moving segments due to the sparser learning signal. Intriguingly, replacing ground truth flow with the one estimated with a state-of-the-art supervised RAFT~\cite{teed2020raft}, or self-supervised SMURF~\cite{stone2021smurf} algorithms barely changes the performance, despite a noticeable decrease in the motion segmentation quality (last column in Figure~\ref{fig:pd}). This result demonstrates the robustness of our method to noise. We use RAFT flow for the remainder of the experiments.

Finally, to better quantify the ability of our model to generalize from sparse, noisy motion segmentations to the whole distribution of objects in crowded scenes, we evaluate the Fg. ARI score of the motion segmentations themselves in the last row of Table~\ref{tab:motion_anal}. We can see that these masks indeed mostly capture the moving objects, however, even for those only a tiny fraction is segmented. In contrast, our approach, capitalizing on this noisy and incomplete signal, increases the overall ARI score by a factor of 15.

\subsection{Comparison to the state of the art}
\label{sec:sota}
Finally, we compare our approach to the state-of-the-art on the validation sets of TRI-PD and KITTI in Table~\ref{tab:sota}. Firstly, we observe that all the learning-based methods fail to achieve non-trivial results on both datasets. This confirms our hypothesis that appearance alone is not a sufficient signal to separate objects from the background in realistic environments. In contrast, our proposed approach outperforms all these methods by a wide margin by capitalizing on independent motion cues.
\begin{table}[bt]
 \centering
  {
\resizebox{\linewidth}{!}{
    \begin{tabular}{l|c@{\hspace{1em}}|c@{\hspace{1em}}|c@{\hspace{1em}}}
    & Learning-based & TRI-PD & KITTI            \\\hline
    SlotAttention~\cite{locatello2020object} & \checkmark & 10.2 & 13.8   \\
    MONet~\cite{burgess2019monet} & \checkmark & 11.0  & 14.9          \\
    SCALOR~\cite{jiang2019scalor} & \checkmark  & 18.6 &  21.1      \\
    S-IODINE~\cite{greff2019multi} & \checkmark & 9.8  &  14.4    \\ 
    MCG~\cite{arbelaez2014multiscale} & \xmark  & 25.1  & 40.9  \\
    \hline
    Ours & \checkmark & \bf{50.9} &  \bf{47.1}  
\end{tabular}
}
}
\caption{Comparison to the state-of-the-art approaches for object discovery on the validation sets of TRI-PD and KITTI using Fg.~ARI. Our approach outperforms both learning- and heuristic-based methods by capitalizing on independent motion cues.}

\vspace{-3mm}
\label{tab:sota}
\end{table}

\begin{figure}[t]
    \centering
    \includegraphics[width = 0.99 \linewidth]{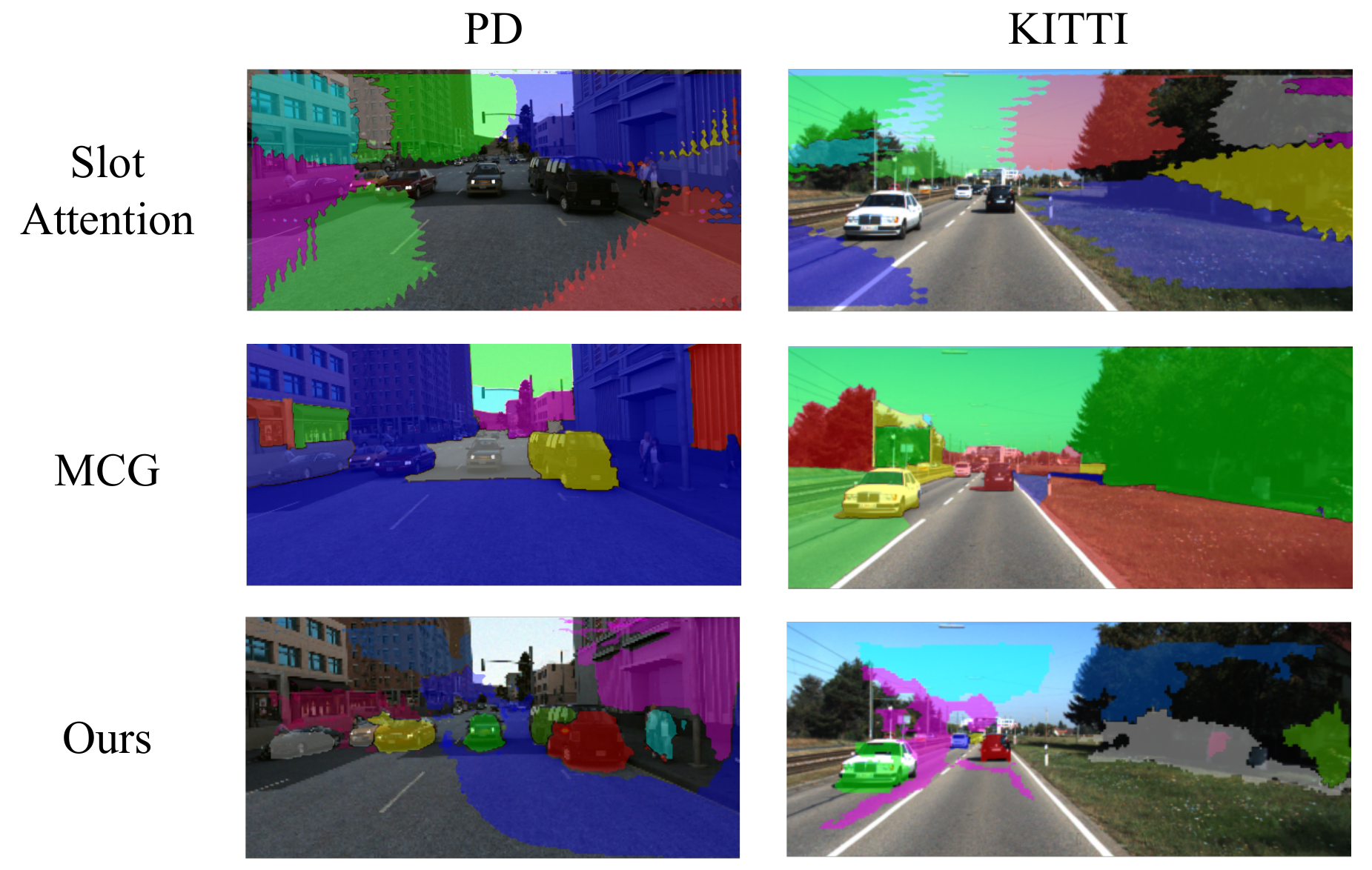}
    \vspace{-2 mm}
    \caption{Qualitative comparison of our approach and representative heuristic- and learning-based methods on the validation sets of TRI-PD and KITTI (showing top-10 masks). Ours learns to successfully separates objects from the background, whereas appearance-based methods struggle in cluttered environments. }
    \vspace*{-4mm}
    \label{fig:kitti}
\end{figure}

Interestingly, the classical MCG approach performs significantly better than the more recent learning-based methods (moreover, this observation holds even on the toy CATER benchmark, as we show in the appendix). Our method outperforms MCG on both datasets, with the margin being significantly larger on TRI-PD. Recall that KITTI is an image-based benchmark, where the annotated frames are selected to prominently feature the objects of interest. In contrast, TRI-PD is a densely labeled video dataset with more challenging camera angles and  more background clutter (see Figure~\ref{fig:kitti} for a qualitative comparison). Thus, wider margins on PD highlight the benefits of our learning-based approach compared to the heuristic-based MCG.

\section{Discussion and limitations}
Discovering objects and their extent from raw data is a challenging problem due to the ambiguity of what constitutes an object. In this work, we propose one way to automatically resolve this ambiguity by focusing on dynamic objects and using independent motion as an inductive bias in an auto-encoder framework. Our analysis demonstrates promising results in real-world environments, while further raising a number of important questions.

\smallsec{Generalization to non-dynamic objects.} While independent object motion provides a convenient signal for object discovery from data, it ignores objects that are not capable of moving by themselves, but might be important for downstream tasks. In particular, in indoor environments people interact with accessories, electronics, food, etc., and capturing these objects is crucial for action recognition~\cite{qi2018learning,zhang2019structured} and robotics~\cite{mousavian20196,billard2019trends}. Notice, however, that extending the definition of a dynamic object to those entities that either move by themselves or can be moved by humans covers most of such cases. Classical motion segmentation approaches~\cite{brox2010object,keuper2015motion} do attempt to capture all the objects that fall into this more general definition, but do not generalize in the wild. Developing more robust, learning-based versions of these methods is a critical step towards a generic object discovery algorithm.

\smallsec{Object category imbalance in the real world.} Like any other learning-based method, ours is susceptible to focusing on the most common categories, while ignoring the objects in the tail of the distribution.
For instance, in the real world we might see lots of moving people, vehicles and animals, and sometimes a person picking up a piece of litter. In theory, this should allow our method to discover not only what people, cars and animals are, but also litter. However, it might happen too infrequently in practice. Fortunately, this problem has received a lot of attention in the few-shot and continual learning domains~\cite{kang2019decoupling,chang2021image,zhang2021videolt,Samuel2021DistributionalRL}, and the proposed solutions can be integrated into our framework.

\smallsec{Supervision used to train the motion segmentation algorithm.} The approach of Dave et al.~\cite{dave2019towards}, used in our experiments, is trained on the toy, synthetic FlyingThings3D~\cite{mayer2016large} dataset with ground truth moving object masks. This raises the question of whether it is this indirect object-level supervision which makes our method outperform other, fully unsupervised approaches. To address this concern, in the supplementary material we directly pre-train SlotAttention on FlyingThings3D in a fully-supervised way, showing this does not have a significant effect on its object discovery performance in realistic videos due to the large domain gap. 

\smallsec{Acknowledgements.}
We thank Alexei Efros, Vitor Guizilini and Jie Li for their valuable comments, and Achal Dave for his help with computing motion segmentations. This research was supported by Toyota Research Institute.

{\small
\bibliographystyle{ieee_fullname}
\bibliography{egbib}
}

\newpage
\appendix
\setcounter{figure}{0}
\setcounter{table}{0}
\renewcommand{\thefigure}{\Alph{figure}}
\renewcommand{\thetable}{\Alph{table}}
\appendix
\appendixpage

In this appendix, we provide additional experimental results, visualizations and implementation details that were not included in the main paper due to space limitations. We begin by analysing several aspects of our approach, including its memory constraints, effect of the network depth, self-supervised backbone pre-training, and influence of the number of slots in Section~\ref{sec:exp}. A separate discussion of our post-processing approach for the motion segmentation outputs, together with a hyper-parameter ablation, is available in Section~\ref{sec:motion}. We then report additional experimental comparisons in Section~\ref{sec:qual} including qualitative comparison with SCALOR, measurement with representative segmentation metrics, and comparison with state-of-the-art on CATER. Finally, we provide more statistics for the synthetic TRI-PD dataset in Section~\ref{sec:pd} and conclude by listing the remaining implementation details in Section~\ref{sec:impl}. 

\section{Further ablations}
\label{sec:exp}

\subsection{Memory constrains}
By adopting the learnable slot initialization and one-shot decoding strategies, our proposed method can greatly save the GPU memory. In Table~\ref{tab:gpu} we compare the memory consumption for both the original SlotAttention architecture~\cite{locatello2020object} and our optimized model on CATER~\cite{girdhar2019cater} and TRI-PD~\cite{parallel_domain} datasets for a single frame. Firstly, we observe that on CATER our approach does results in an about 25\% reduction in the amount of memory required to train the model, though both methods are easy to fit on a single GPU due to the low resolution of CATER frames and a small number of slots. In contrast, on TRI-PD, where both the resolution and the number of slots are much larger, the memory constraints of SlotAttention become prohibitive whereas our proposed architecture can save 90\% of the GPU memory, enabling experiments on this realistic dataset.

\subsection{Deeper backbones}

\begin{figure}
    \centering
    \includegraphics[width = \linewidth]{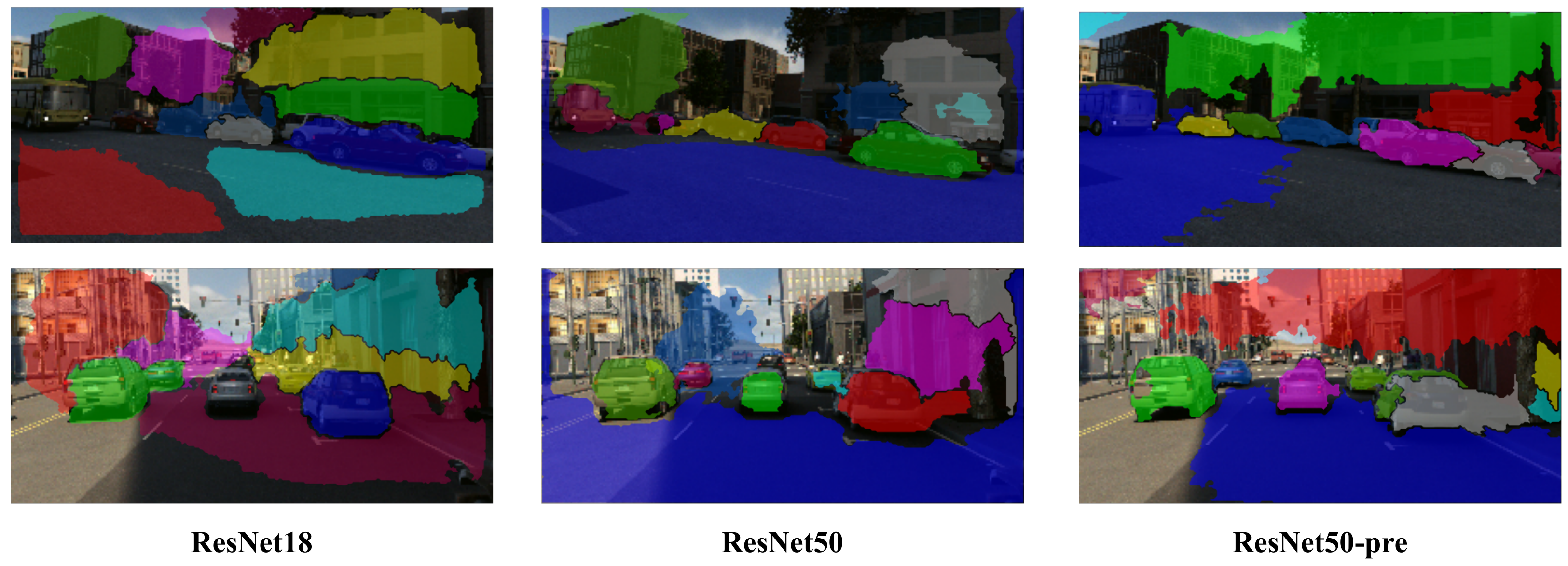}
    \caption{Visualizations of two samples on TRI-PD dataset with different backbones. Deeper backbone results in a higher confidence for the foreground objects, and self-supervised pre-training of the backbone helps to better capture the object masks.}
    \label{fig:backbone}
\end{figure}

In the main paper, we used a ResNet18 backbone for all the experiments. We now further evaluate the proposed approach with ResNet34 and ResNet50 backbones on the TRI-PD dataset in Table~\ref{tab:backbone}. In addition, we explore the effect of self-supervised ImageNet~\cite{deng2009imagenet} pre-training of the backbone with the recent SimSam~\cite{chen2021exploring} approach. We evaluate these variants with both ground truth motion segmentation and the outputs of~\cite{dave2019towards} with RAFT flow, reporting Fg.~ARI for both static and moving instance. We also visualize two generated samples for randomly initialized ResNet50 and pre-trained ResNet50, together with our ResNet18 baseline, for RAFT + \cite{dave2019towards} setting in Figure~\ref{fig:backbone}. 

Firstly, we observe that increasing the network depth indeed results in consistent performance improvements of our approach both with GT and estimated motion segmentation, but the improvements are somewhat larger in the former setting. This shows that the noisy estimated motion segmentation limits the performance of our method and improvements in motion segmentation algorithms would directly result in a better scalability. Secondly, self-supervised pre-training of the backbone results in further improvements for both variants, demonstrating that recent advances in self-supervised representation learning can be easily combined with our object discovery approach.

\begin{table}[bt]
 \centering
  {
\resizebox{\linewidth}{!}{
\begin{tabular}{l|l|c|c|l}
Model & Dataset & Resolution & \#Slots & GPU Memory      \\\hline
    SlotAttention~\cite{locatello2020object} & CATER & $128 \times 128$ & 10 & 652 MB \\
    Ours & CATER & $128 \times 128$ & 10 & 483 MB \\
    SlotAttention~\cite{locatello2020object} & TRI-PD & $548 \times 1123$ & 45 & 23,796 MB \\
    Ours & TRI-PD & $548 \times 1123$ & 45 & 2,297 MB  \\
\end{tabular}
}
}
\caption{GPU memory consumption for SlotAttention and our proposed method measured with Megabytes (MB). By applying learnable slot initialization and one-shot decoding, our model can save 90\% of the GPU memory on the realistic TRI-PD datset with a large resolution and dozens of objects.}
\label{tab:gpu}
\end{table}

\begin{table}[bt]
 \centering
  {
\resizebox{\linewidth}{!}{
    \begin{tabular}{l|l|c@{\hspace{1em}}|c@{\hspace{1em}}|c@{\hspace{1em}}}
    Backbone & Motion seg. & Fg. ARI Stat. & Fg. ARI Mov. & Fg. ARI All            \\\hline
    ResNet18& GT moving & 48.4 & 66.7 & 59.6 \\
    ResNet18 & RAFT flow + \cite{dave2019towards}  & 45.6 & 56.7 & 50.9  \\
    ResNet34 & GT moving & 50.1 & 69.0  & 61.3 \\
    ResNet34 & RAFT flow + \cite{dave2019towards}  & 46.2 & 57.1 & 51.7  \\
    ResNet50 & GT moving & 51.3  & 69.7 & 62.0\\
    ResNet50 & RAFT flow + \cite{dave2019towards}  & 47.2 & 57.2 & 52.0  \\ \hline
    ResNet50-pre & GT moving & \bf 53.6 & \bf  71.2  & \bf 64.1 \\
    ResNet50-pre & RAFT flow + \cite{dave2019towards}  & 48.5 & 68.6 & 53.1  \\
\end{tabular}
}
}
\caption{Analysis of the effect of the backbone depth and self-supervised pre-training on the model's performance on the validation set of TRI-PD. Both deeper backbones and better weight initialization result in performance improvements with either GT or estimated motion segmentation, but the improvements are somewhat higher in the former setting.
}
\vspace{-4mm}
\label{tab:backbone}
\end{table}

\subsection{Stronger SlotAttention baselines}
To further validate the effectiveness of both the proposed architecture and motion supervision, we now report two additional baseline for SlotAttention~\cite{locatello2020object}. Firstly, we apply the same learning signal in the form of motion masks to~\cite{locatello2020object} on CATER and report the results in Table~\ref{tab:slot-cater}. We observe that the independent motion prior indeed also improves the performance of the SlotAttention, but it remains 9.6 Fg.~ARI points below our method. This result indicates that our model architecture not only dramatically reduces memory consumption, enabling the experiments on the realistic TRI-PD and KITTI datasets, but also improves the object discovery capabilities of the approach. 

\begin{table}[bt]
 \centering
  {
\resizebox{\linewidth}{!}{
    \begin{tabular}{l|c|c}
    Model & Motion masks  & Fg. ARI            \\\hline
    SlotAttention & \xmark  & 64.4 \\
    SlotAttention & \checkmark & 83.1 \\ 
    Ours & \checkmark  & \bf 92.7 \\
\end{tabular}
}
}
\caption{Comparison with SlotAttention using motion masks supervision on CATER. Independent motion signal can also improve the perofrmance of this baseline, but it remains below that of our model, indicating the effectiveness of our model design.
}
\vspace{-4mm}
\label{tab:slot-cater}
\end{table}

Next, we explore whether the supervised pre-training of the motion segmentation approach of Dave et al.~\cite{dave2019towards} on the FlyingThings3D dataset~\cite{mayer2016large} provides an unfair advantage to our method. To this end, we pre-train SlotAttention on FlyingThings3D, which is a more direct form of utilizing these labels, and report the results in Table~\ref{tab:slot-pd}. The results indicate that while pre-training on object segmentation labels in~\cite{mayer2016large} does result in a small improvement for the SlotAttention, its the performance remains low and the model fails to discover the objects in realistic images. This is due to the large domain gap between the toy FlyingThings3D and the photo-realistic TRI-PD datasets. This toy data, however, is sufficient to learn to segment moving objects in the optical flow field - a low level task with an appearance agnostic input. The resulting model can then be used to bootstrap object discovery in real world environments.

\begin{table}[bt]
 \centering
  {
\resizebox{\linewidth}{!}{
    \begin{tabular}{l|l|c}
    Model & Pre-training  & Fg. ARI            \\\hline
    Slot Attention &  None & 10.2 \\
    Slot Attention & FlyingThings3D~\cite{mayer2016large} & 19.1 \\ 
    Ours & None  & \bf 50.9 \\
\end{tabular}
}
}
\caption{Comparison with SlotAttantion pre-trained on FlyingThings3D~\cite{mayer2016large} on the validation set of TRI-PD. Direct pre-training only results in minor improvements for SlotAttention, whereas using these labels to train a motion segmentation approach which later bootstraps object discovery in our framework is a much more effective strategy. 
}
\vspace{-4mm}
\label{tab:slot-pd}
\end{table}

\subsection{Influence of the number of slots}
\begin{figure}[ht]
    \centering
    \vspace{-5 pt}
    \includegraphics[width = 0.9 \linewidth]{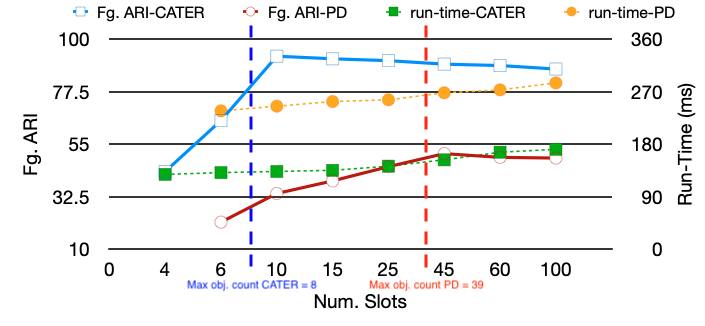}
    \vspace{-10 pt}
    \caption{Influence of the number of slots on our method's performance and run-time on CATER and TRI-PD. }
    \label{fig:curve}
    \vspace{-4mm}
\end{figure}

In the main paper, we use a fixed number of slots, which is slightly larger than the maximum number of objects for each dataset. Here we ablate the effect of the number of slots on our method's performance and run-time on CATER and TRI-PD in Figure~\ref{fig:curve}. Firstly, we observe that, unsurprisingly, using fewer slots than the maximum object count in a dataset results in a decrease in performance. However, increasing the slot number has a minimal effect on Fg.~ARI and run-time. The latter is due to our efficient 1-shot decoding strategy, described in Section~\ref{sec:scale} of the paper. These results demonstrate the flexibility of our method, which does not require the ground truth object count for training.

\section{Motion-segmentation post-processing}
\label{sec:motion}
We now describe the motion segmentation post-processing steps applied in our work. Firstly, we filter out extremely small (fewer than 100 pixels, or smallest dimension of the enclosing bounding box less than 10) and extremely large (occupying more than 60\% of the image) segments since those typically correspond to random noise or capture background regions. We then additionally remove the segments that are within 15 pixels from the image boundary, as well as segments containing more than one connected component, which also typically correspond to background and noisy regions respectively. 

These rules are applied to the outputs of both the heuristic-based (CUT)~\cite{keuper2015motion} and the learning-based (TSAM)~\cite{dave2019towards} motion segmentation algorithms, with the only difference being that, since~\cite{keuper2015motion} outputs spatio-temporal segments, we average the frame-level statistics over time. For the method of Dave et al.~\cite{dave2019towards} we directly apply the rules at every frame.  

In addition, unlike the heuristic-based approach, the method of Dave et al.~\cite{dave2019towards} also predicts a confidence score for each segment and applies an internal pre-processing step to the optical flow, zeroing out flow vectors with a low magnitude, since those are unreliable. We integrate both of these components into our post-processing algorithm by filtering out segments with confidence score lower than $T_{conf}$, and average normalized flow magnitude lower than $T_{mag}$ (we first normalize flow magnitude to be $\in [0, 1]$ for each frame).

\begin{figure*}
    \centering
    \includegraphics[width = \linewidth]{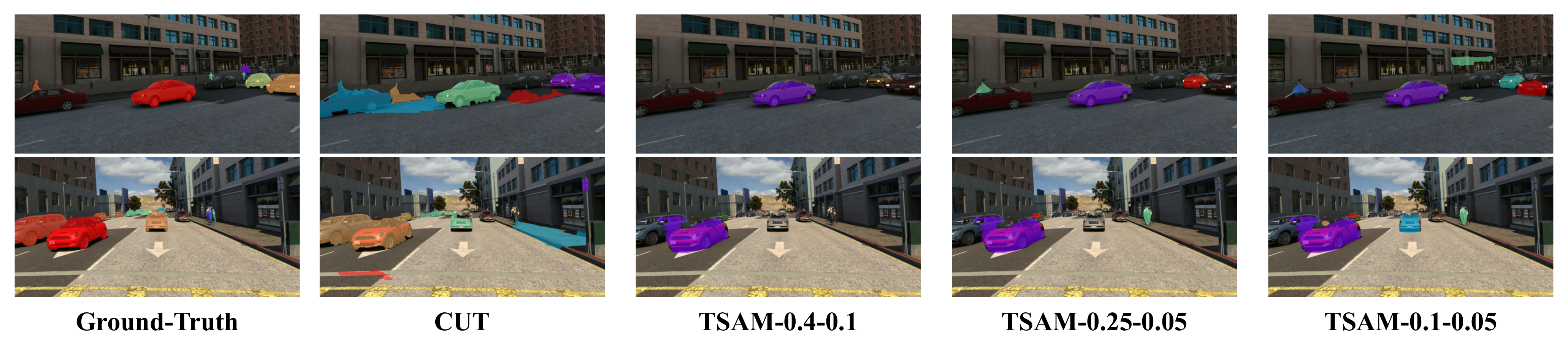}
    \caption{Visualizations of the motion segmentation post-processing with different methods and thresholds. For TSAM, with a loosing constrain, we can find more segments, but also more noisy parts. We set $T_{con}$ as 0.25 and $T_{mag}$ as 0.05 to balance the quality and quantity of the segments.}
    \label{fig:motion}
\end{figure*}

We select these two thresholds on the validation set of TRI-PD using the FG.ARI score between the post-processed motion segments and ground-truth segments corresponding to the moving objects in Table~\ref{tab:motion}. In addition, we report the average number of segments per frame after post-processing under $\#Seg$. Firstly, we evaluate the motion segments produced by~\cite{keuper2015motion} for reference and observe that while this approach outputs more segments, its accuracy is quite low, as indicated by the mIoU score. In contrast, the learning-based approach of Dave et al.~\cite{dave2019towards} produces fewer segments, but they are a lot more accurate across a variety of thresholds. We also visualize 2 sample frames in Figure~\ref{fig:motion} for a qualitative comparison. For our main paper, we set $T_{con}$ to 0.25 and $T_{mag}$ to 0.05 to balance segmentation precision and recall.

\begin{table}[t]
    \centering
    \resizebox{\linewidth}{!}{
    \begin{tabular}{l|cc|c|c}
          Method & $T_{con}$ & $T_{mag}$ & \#Seg &  mIoU ($\uparrow$) \\ \hline
      CUT~\cite{keuper2015motion} & - & - & 1.53 & 2.5 \\ \hline
        TSAM~\cite{dave2019towards} & 0.4 & 0.1 & 0.38 & 2.9 \\
        TSAM~\cite{dave2019towards} & 0.4 & 0.05 & 0.51 & 3.1 \\
        TSAM~\cite{dave2019towards} & 0.25 & 0.1 & 0.57 & 3.2 \\
      TSAM~\cite{dave2019towards} & 0.25 & 0.05 & 0.63 & 3.4 \\
      TSAM~\cite{dave2019towards} & 0.1 & 0.1 & 0.71 & 3.4 \\
      TSAM~\cite{dave2019towards} & 0.1 & 0.05 & 0.92 & 3.3 \\
    \end{tabular}
    }
    \caption{Fg.~ARI measurements and averaged number of segments on TRI-PD dataset with different threshold for the post-processed motion segments. We set $T_{con}$ as 0.25 and $T_{mag}$ as 0.05 in the main paper based on the ARI scores.}
    \label{tab:motion}
\end{table}

\section{Additional experimental comparisons}
\label{sec:qual}
\subsection{Qualitative comparison to SCALOR}

\begin{figure}
    \centering
    \includegraphics[width = \linewidth]{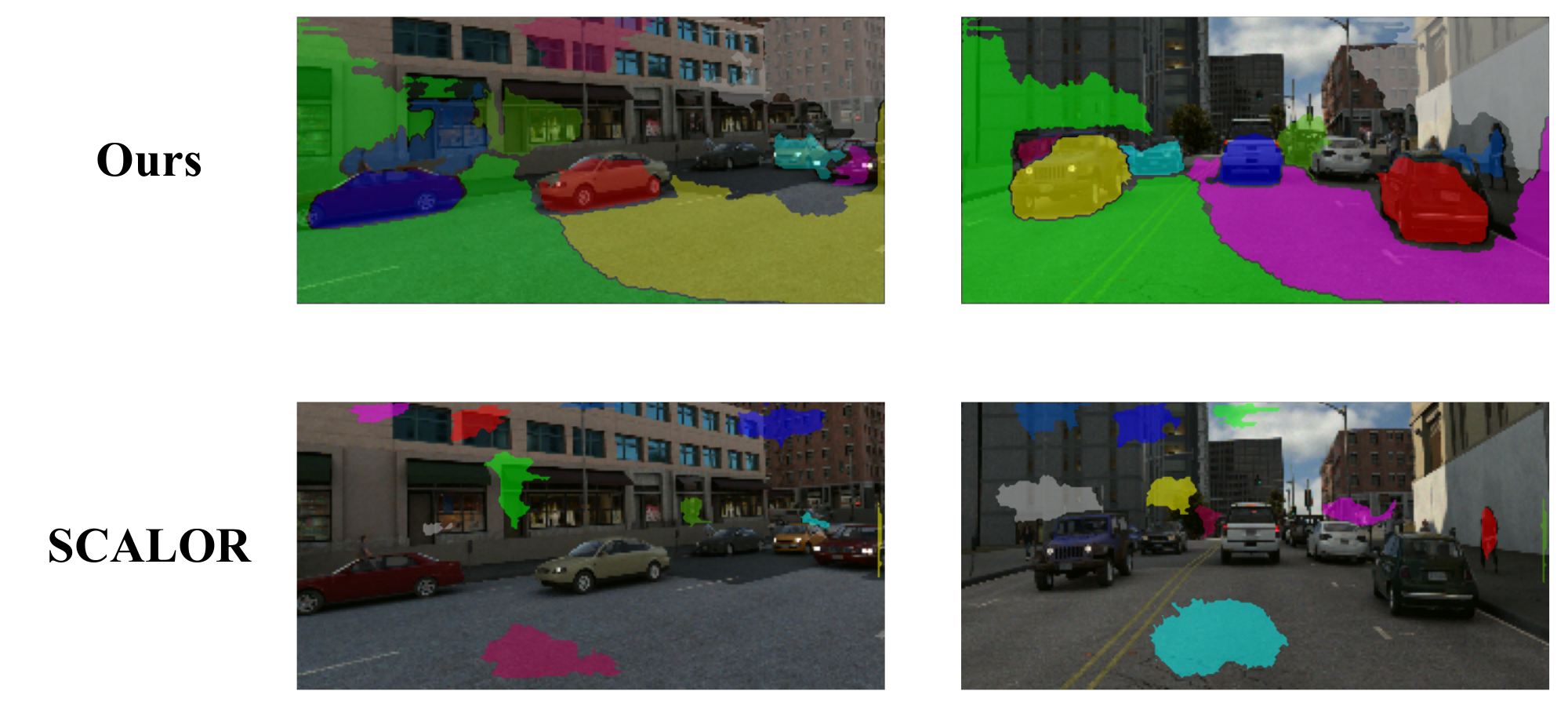}
    \caption{Visual comparison of our method and SCALOR. Despite relatively high performance, this approach also fails to discover most of the objects. }
    \label{fig:scalor}
\end{figure}

Here we qualitatively compare our approach to the top-preforming SCALOR~\cite{jiang2019scalor} baseline on the validation set of TRI-PD in Figure~\ref{fig:scalor}. Since SCALOR does not provide scores for the generated segments, we sampled 10 masks uniformly to generate the visualizations. The results indicate that SCALOR also did not work well with complicated backgrounds and could not discover the objects. 

\subsection{Evaluation with segmentation metrics}
We use Fg.~ARI as the standard metric for object discovery in the main paper. The key reason is that it ignores the (unimportant) differences in how methods segment the background. For a more comprehensive understanding of the method, we also evaluate with  F-measure~\cite{ochs2013segmentation} and mIoU, which are more standard segmentation metrics, on TRI-PD dataset and report the results in Table~\ref{tab:fmeasure}. Firstly, we observe that our method still outperforms prior work on both metrics. Secondly we notice that F-measure has the same property as Fg.~ARI (ignoring the background segments) and provides similar conclusions. In contrast, mIoU penalizes background over-segmentation and thus is less informative in this setting.

\begin{table}[t]
    \centering
    \vspace{-5 pt}
    \resizebox{ \linewidth}{!}{
    \begin{tabular}{c|ccccc|c}
    & Slot Attention & MONet & SCALOR & S-IODINE & MCG & Ours \\ \hline 
    Fg.~ARI & 10.2 & 11.0 & 18.6 & 9.8 & 25.1 & {\bf 50.9} \\
    F-measure & 11.0 & 9.4 & 14.1 & 10.2 & 25.8 & {\bf 47.1} \\
    mIoU & 9.2 & 7.7 & 12.9 & 13.6 & 24.5 & {\bf 38.0} \\
    \end{tabular}
    }
    \vspace{-10 pt}
    \caption{Evaluation on TRI-PD with with Fg.~ARI, F-measure and mIoU. Metrics that do not penalize background over-segmentation are more informative but our approach shows top results overall.}
    \label{tab:fmeasure}
    \vspace{-10 pt}
\end{table}

\subsection{Comparison with SOTA on CATER}

For completeness, we now compare our approach (with estimated motion segmentation) to state-of-the-art on the toy CATER dataset, and report the results in Table~\ref{tab:cater}. For these experiments, we use the original, shallow backbones for prior works, in contrast to ResNet18 used in KITTI evaluation, since we observed that they achieve higher performance on CATER. We find that: (1) the baselines' performances are lower compared to CLEVR used in these works due to higher scenes complexity (\eg, more occlusions); (2) The conclusions from the main paper hold, with our method showing top results; (3) heuristic-based MCG outperforms most of the recent object-centric learning approaches even on this toy dataset, highlighting the importance of using strong baselines. 

\begin{table}[ht]
    \centering
    \vspace{-5 pt}
    \resizebox{ \linewidth}{!}{
    \begin{tabular}{c|ccccc|c}
         & SlotAttention & MONet & S-IODINE & SCALOR & MCG & Ours \\ \hline 
        Fg.ARI & 67.3 & 88.6 & 73.5 & 74.6 & 84.0 & \bf 90.4 \\
        
    \end{tabular}
    }
    \vspace{-10 pt}
    \caption{Comparison with prior art on CATER. Our method shows top results, and MCG outperforms most learning-based methods.}
    \label{tab:cater}
    \vspace{-10 pt}
\end{table}

\section{Parallel Domain dataset details}
\label{sec:pd}

\begin{figure*}[ht]
    \centering
    \includegraphics[width = 0.95 \linewidth]{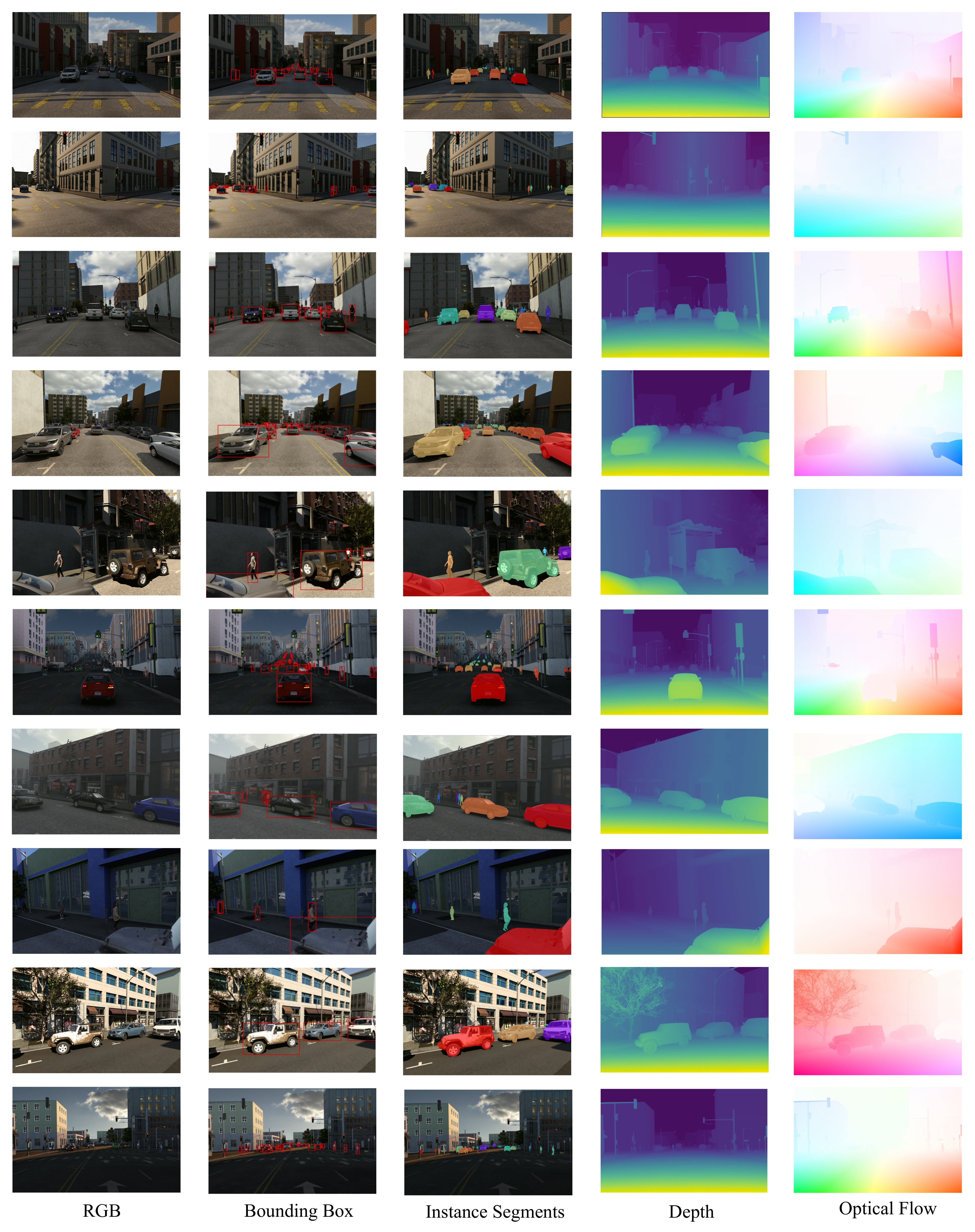}
    \caption{Several samples from the synthetic TRI-PD dataset, together with corresponding annotations.}
    \label{suppfig:pd}
\end{figure*}

In this section we describe the details of our synthetic TRI-PD dataset, which was collected through a state-of-the-art synthetic data generation service~\cite{parallel_domain}. The whole dataset contains 200 photo-realistic scenes with driving scenarios in city environments captured at 20 FPS. Each video is 10 seconds long, with a fixed shape at $1936 \times 1216$, and comes with 7 different independent camera views. A comprehensive set of ground truth labels is provided for every video, which include: camera pose, calibration, depth, instance segmentation, semantic segmentation, 2D bounding box, 3D bounding box, depth, forward 2D motion vectors, backward 2D motion vectors, forward 3D motion vectors, and backward 3D motion vectors. Figure~\ref{suppfig:pd} shows several samples of the data and corresponding annotations. We filter out scenarios with low visibility (e.g. foggy and dark scenes) which are not useful in the context of object discovery, resulting in 154 scenes which we use for training. In addition, we render another 17 scenes separately for evaluation. We use 6 camera views for training, and 3 for evaluation, resulting in 924 training and 51 test videos. The dataset is available at our project web page: \url{https://github.com/zpbao/Discovery_Obj_Move}.

We define the objects belonging to the following categories as the foreground objects: Pedestrian, Bus, Car, Bicyclist, Caravan/RV, OtherMoveable, Motorcycle, Motorcyclist, OtherRider, Train, Truck, ConstructionVehicle, and Bicycle. We filter out the objects with over 50\% occlusion and fewer than 150 visible pixels. To find the independently moving objects in a pair of consecutive frames $F^t, F^{t+1}$, we first propagate all the object centers from $F^t$ to $F^{t+1}$ with the ground truth camera motion. Then we calculate the distances between these propagated object centers and the ground truth centers in $F^{t+1}$. If the distance is larger than 0.05, we label that object as moving independently from the camera.

We also report some statistics for each foreground category, including averaged number of object per frame (Num./f), Ratio of Moving objects (RoM), Ratio of Static objects (RoS), Small Object ratio (SObj), Medium Object ratio (MObj), and Large Object ratio (LObj) in Table~\ref{tab:pd}. Notice that for some foreground categories, there is no object found in our subset. We define objects that cover fewer than 2000 pixels in the original resolution as small objects, larger than 2000 pixels but fewer than 30000 pixels as medium objects, and larger than 30000 pixels as large objects. We only count the objects for which at least 50\% of the object mask is visible. Although for most categories more than half of the objects are in motion, in practice only a tiny fraction of these objects are captured by our motion segmentation algorithm (see low FG.ARI values in Table~\ref{tab:motion}).

\begin{table}[t]
    \centering
    \resizebox{\linewidth}{!}{
    \begin{tabular}{l|cccccc}
          Category & Num./f & RoM & RoS & SObj & MObj & LObj \\ \hline
          Pedestrian & 0.12 & 0.76 & 0.24 & 0.82 & 0.16 & 0.02\\ 
          Bus & 0.13 & 0.73 &  0.27 & 0.21 & 0.46 & 0.33 \\ 
          Car & 5.44 & 0.37 & 0.63 & 0.49 & 0.38 & 0.13\\
          Bicyclist & 0.15 & 0.73 & 0.27 & 0.79 & 0.19 & 0.02 \\
          Caravan/RV & 0.05 & 0.75 & 0.25 & 0.20 & 0.56 & 0.24 \\
          OtherRider & 0.09 & 0.78 & 0.22 & 0.72 & 0.25 & 0.03\\
          ConstructionVehicle & 0.01 & 0.78 & 0.22 & 0.72 & 0.25 & 0.03\\
          Bicycle & 0.15 & 0.73 & 0.27 & 0.79 & 0.19 & 0.02\\
       
    \end{tabular}
    }
    \caption{Statistics of Parallel Domain (TRI-PD) dataset. The averaged number of object per frame (Num./f), Ratio of Moving objects (RoM), Ratio of Static objects (RoS), Small Object ratio (SObj), Medium Object ratio (MObj), and Large Object ratio (LObj) are reported.}
    \label{tab:pd}
\end{table}

\section{Implementation details}
\label{sec:impl}

To increase the output resolution of the feature map of the encoder, we modify the standard PyTorch implementation of a ResNet~\footnote{\url{https://github.com/pytorch/vision/blob/main/torchvision/models/resnet.py}}. In particular, we reduce the downsampling ratio from 16 to 4 by using stride 1 for for all the convolutional blocks except for the first one. We further drop the last fully-connected layers of the ResNet to obtain a feature map. For the decoder, we adopt the 4-layer shallow decoder following~\cite{locatello2020object}.

All the models are trained for 500 epochs using Adam~\cite{kingma2014adam} with a batch size 20 and learning rate 0.001. Following~\cite{locatello2020object}, we use a learning rate warm up for 2000 iterations. For the exponential learning rate decay schedule, we set the decay rate as 0.5 and the decay step as 500000. We set $\lambda_M$ to 0.5 and $\lambda_T$ to 0.01. $D_{slot}$ and the output dimension for convGRU are set to 64. 

To convert the attention maps $W$ to segmentation masks, we first apply a SoftMax along the slot dimension to obtain a distribution over slots for each pixel. We then take the argmax of this distribution to assign each pixel to one of the slots and treat the resulting assignment as the masks.

\end{document}